\documentclass{article}

\usepackage[final]{neurips_2024}

\usepackage{titlesec}

\usepackage[utf8]{inputenc} 
\usepackage[T1]{fontenc}    
\usepackage{hyperref}       
\usepackage{url}            
\usepackage{booktabs}       
\usepackage{amsfonts}       
\usepackage{nicefrac}       
\usepackage{microtype}      
\usepackage{subcaption}
\usepackage{enumitem}
\usepackage{amsmath}
\usepackage{graphicx}
\usepackage[table]{xcolor}  
\newcommand{\highlight}[1]{\cellcolor{yellow!25}\textbf{#1}}

\title{Air in Your Neighborhood: Fine-Grained AQI Forecasting Using Mobile Sensor Data}

\author{
	Aaryam Sharma \\
	School of Computer Science\\
	University of Waterloo\\
	Waterloo, ON, N2L 3G1 \\
	\texttt{a584shar@uwaterloo.ca} \\
}

\begin{document}
\maketitle

\begin{abstract} 

Air pollution has become a significant health risk in developing countries. While governments routinely publish air-quality index (AQI) data to track pollution, these values fail to capture the local reality, as sensors are often very sparse. In this paper, we address this gap by predicting AQI in 1 km$^2$ neighborhoods, using the example of AirDelhi dataset. Using Spatio-temporal GNNs we surpass existing works by 71.654 MSE a 79\% reduction, even on unseen coordinates. New insights about AQI such as the existence of strong repetitive short-term patterns and changing spatial relations are also discovered. The code is available at \url{https://github.com/ASChampOmega/AQI_Forecasting.git}

\end{abstract} 

\vspace{-5pt}
\section{Introduction}
\vspace{-5pt}

Accurate prediction of Air Quality Index (AQI) in cities like New Delhi is a pressing need because of the health impacts of air pollution. In New Delhi, the AQI levels are often hazardous in the winter and has caused several respiratory and cardiovascular diseases. According to recent studies, the air pollution problem in New Delhi causes an estimated 10,000 premature deaths yearly \citep{chen_air_2025}. Economically, there have also been significant impacts - air pollution resulted in losses of \$36.8 billion in 2019, equivalent to 1.36\% of India's GDP \citep{WorldBank} have been accrued.

Particulate matter in particular presents a major threat to health because our breathing cannot filter these particles out. \citet{sahu_validation_2020} recommend placing one sensor per square kilometer. However installing such a dense network is not only expensive, they only capture current conditions and do not provide insight into the future, limiting their use in preventive action.

\begin{figure*}[]
    \centering
    \begin{subfigure}{0.3\textwidth} 
        \centering
        \includegraphics[width=\linewidth]{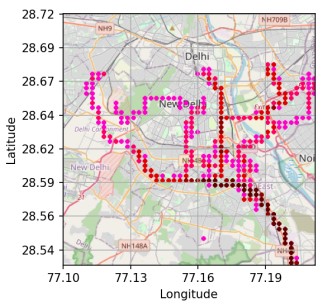} ~~~
        \caption{AirDelhi Sampled points Heatmap - darker colours indicate more points \citep{AirDelhi}}
        \label{fig:bus_routes}
    \end{subfigure}
    \begin{subfigure}{0.3\textwidth}
        \centering
        \includegraphics[width=\linewidth]{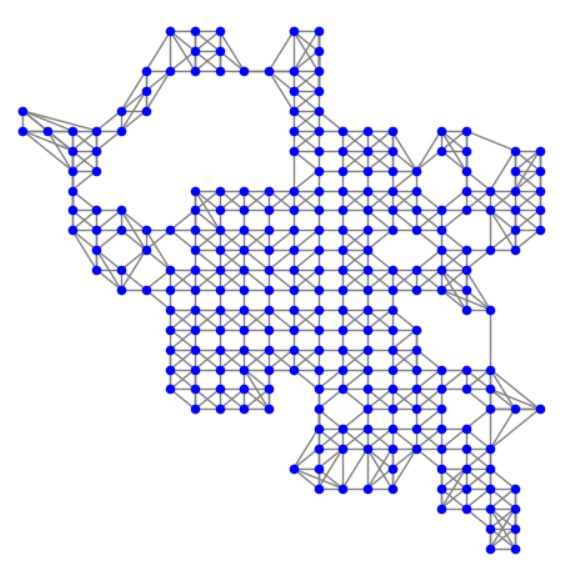} ~~~
        \caption{Grid generated by bucketing bus routes and the 5 nearest-neighbor graph for the "Extended" evaluation dataset}
        \label{fig:gnn_graph}
    \end{subfigure}
    \begin{subfigure}{0.3\textwidth} 
        \centering
        \includegraphics[width=\linewidth]{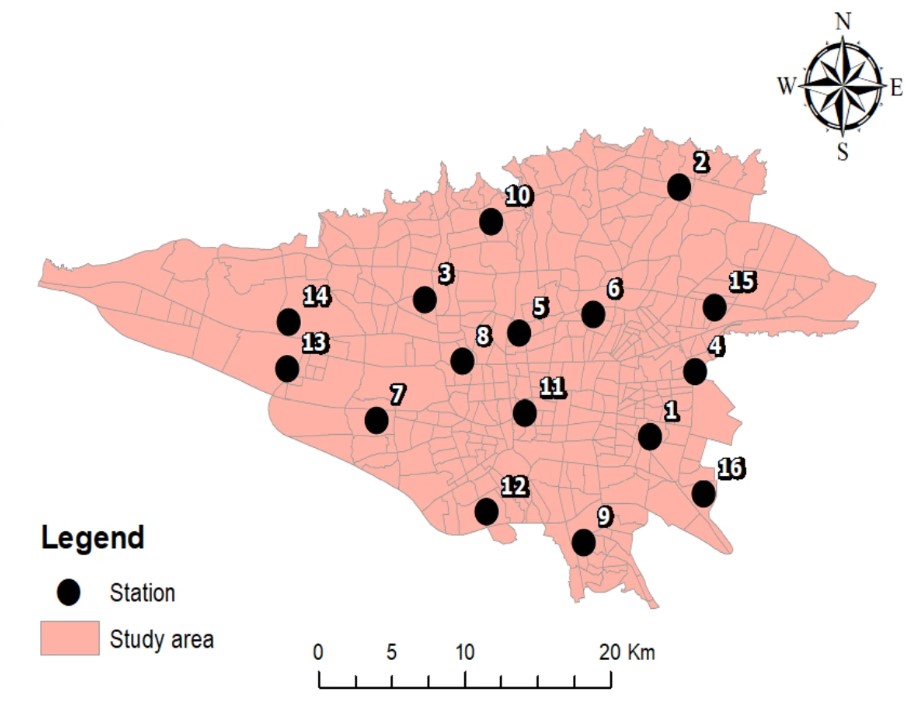} ~~~
        \caption{Map of data from Tehran AQI Prediction \citep{rad_predictive_2025}} The sensors are only a few in number.
        \label{fig:sensors}
    \end{subfigure}
    \caption{Figures show the different nature of data. Bus routes are continuous paths, whereas typical research occurs on sparse stations as in figure (c). There are still gaps in the data in (a).The cell based discretized representation in (b) has more points than (a), because several cells are not visited daily.}
    \label{fig:eqv_sde_traj}
    \vspace{-15pt}
\end{figure*}

This research predicts the PM$_{2.5}$ and PM$_{10}$ measurements at a fine-grained level, which would allow governments to implement solutions at a localized level and reduce spending on costly physical sensors. The primary data source in this paper is the AirDelhi dataset, which was collected by fitting mobile-sensors on buses (Figure \ref{fig:bus_routes}), covering a large area spanning 559 km$^2$ \citep{AirDelhi}. 

Fine-grained AQI prediction is challenging, especially for New Delhi for several reasons, including the large scale of the AirDelhi dataset, higher variance, and presence of unique statistical characteristics compared to previous significantly smaller studies \citep{AirDelhi}. Furthermore, mobile sampling is highly irregular, which introduces data processing challenges as attempting to extract time series for any coordinate results in mostly empty sequences. From a modeling perspective, this is a unique forecasting problem because there are both spatial and temporal characteristics that depend on several factors for which data is scarce and often restricted.

The main contributions of this report are:
\begin{itemize}[itemsep=-1pt, topsep=0pt]
    \item \textbf{First deep learning centric study} on the large-scale mobile sensing dataset AirDelhi.
    \item \textbf{Significantly improve the performance} of 24-hour AQI forecasting by reducing MSE from 90.6 to 18.96 using models such as GRUs and Spatio-Temporal GNNs.
    \item \textbf{Present new insights about the dataset} by conducting several ablation studies revealing the existence of repetitive short-term temporal patterns and changing spatial relations.
    \item \textbf{Provide a practical modeling framework} for mobile sensor data. The framework can also integrate satellite imagery into a temporal dataset.
\end{itemize}

\vspace{-5pt}
\section{Related Works}
\vspace{-2pt}
Due to the severe economic and health risks, the study of AQI prediction has been conducted across the globe. Historically, most of the AQI prediction research is done at the station/sensor level. These stations and sensors are often sparsely scattered due to their high cost (Figure \ref{fig:sensors}). This data is often private and limits future research and reproducibility. Most of my reviewed papers use a different dataset, even for the same city \citep{AirDelhi, pruthi_low-cost_2022, alawi_temporal_2024}.

From the modeling perspective, the majority of research has been done using statistical methods such as Vector Linear Regression, Decision Trees, Random Forests, Gaussian Process, and ensembles. Recently, many research papers are using deep learning methods including LSTMs and GRUs \citep{mendez_machine_2023, parameshachari_prediction_2022}. Many modeling approaches also include external data sources such as wind, pressure, and satellite imagery, from which feature extraction is performed using CNNs. The combination of these features across is often done using Graph Neural Networks and Graph attention networks \citep{mendez_machine_2023, hettige_airphynet_2024, guan_fine-scale_2020, zhao_astgc_2023, rowley_predicting_2023}.

However, there has been limited research from a fine-grained perspective. Most papers focus on economizing low-cost networks to measure AQI in real-time instead of fine-grained forecasting. We were only able to find one similar paper to this proposal, Deep-MAPS, which uses several modalities of data such as weather, population, geography to augment their model. However, they do not use deep learning, the data is inaccessible, and they do not produce forecasts \citep{song_deep-maps_2020}. The AirDelhi paper also only focuses on classical methods, and the benchmarks (most of which we replicated in Table \ref{tab:model_results}) they provide no feature engineering or use of spatio-temporal relations except in their GraphSage model, which does not have a strong performance \citep{AirDelhi}.
\vspace{-5pt}
\section{Problem Formulation and Evaluation}
\vspace{-2pt}
The AQI forecasting task is in a formulation, typical of time series forecasting problems. For the AirDelhi dataset, the original paper aims to make a forecast for 24 hours \citep{AirDelhi}. This is the main target for this paper. The ablation studies explore longer forecasts.

The evaluation metrics are the Mean Squared Error (MSE), Mean Absolute Error (MAE), and $R^2$ \citep{AirDelhi, nguyen_predicting_2024, natarajan_optimized_2024}. $R^2$ score is $1 - \frac{RSS}{TSS}$ where RSS represents the residual sum of squares and TSS represents the total sum of squares of the data. A negative $R^2$ indicates performance worse than the mean, while perfect predictions have an $R^2$ of 1.

I will also report on the problem as a classification task based on the categories outlined by \citep{India_AQI} for dangerous AQI. The categories are based on pollutant thresholds: Good (PM$_{10}$: 0–50, PM$_{2.5}$: 0–30), Satisfactory (51–100, 31–60), Moderately Polluted (101–250, 61–90), Poor (251–350, 91–120), Very Poor (351–430, 121–250), and Severe (PM$_{10}$ >430, PM$_{2.5}$ >250).

Based on \citet{sahu_validation_2020}, the data is divided into grids of size 1 square km. This also maintains a reasonable amount of data per coordinate and improves problem tractability. Measurements are further grouped into 30-minute intervals. Any readings outside 5:30 AM IST to 10 PM IST are discarded. This processing is the same as \citep{AirDelhi}. In addition, several grid cells have fewer than 2 readings per day which were discarded as outliers. These were used as an "Extended" evaluation set to test model performance on unseen coordinates (See Table \ref{tab:model_results}).

The data is split across time as an 80-20 train-test split. None of the test data is available during training. For every training or test sample, the features are only from the past, and any future data leakage has been prevented to ensure the validity of results.

\vspace{-5pt}
\section{Modeling}
\vspace{-2pt}
\subsection{Data Pre-processing}

AQI prediction depends on various factors including: air pressure, wind, pollution in neighboring cities, traffic, foliage, type of neighborhood, population density, and more. While the scope of this research is limited to AirDelhi and publicly available data, several factors can be modeled softly. 

The features available in AirDelhi are: datetime, longitude, latitude, PM$_{2.5}$ and PM$_{10}$ concentration, and device ID. To model traffic softly, the average speed of the bus in the 30-minute window of each spatiotemporal cell (ST-cell) can be calculated. The number of distinct devices (bus count) in a cell also informs us about the traffic in the area. 7-day lagged measurements of PM$_{2.5}$ and PM$_{10}$ are used to capture long-term relations. To model foliage, neighborhood type, and population, ESRI's satellite images are used. Features from these images are extracted using a CNN.

\begin{figure}[htbp]
    \hspace{-1cm}
    \includegraphics[width=1.2\textwidth]{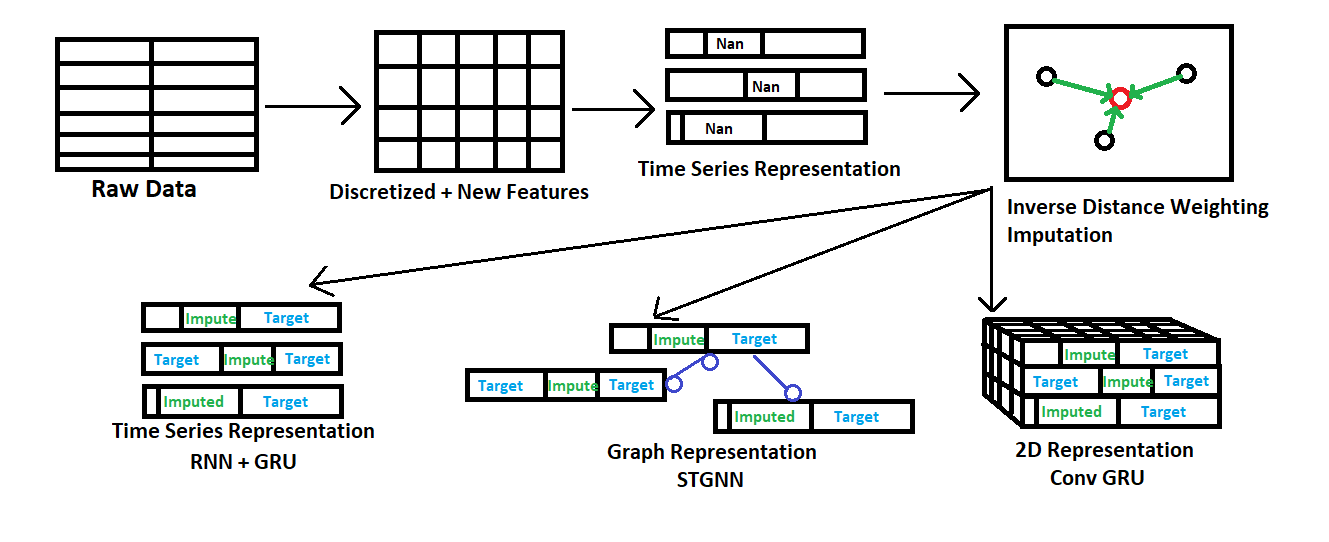}
    \caption{An overview of the feature engineering pipeline. IDW is used for imputation. Three primary final representations are used: Time Series, Graph and 2D and the models they are used for are listed. Any loss is backpropagated only over true ground truth labels ("Target" above)}
    \label{fig:feat_eng}
\end{figure}

The data is missing several entries (when there is no bus with a sensor traveling through a ST-cell) despite removing infrequent coordinates. To have a complete time series as input to the RNNs and GRUs, an Inverse Distance Weighting model, with the closest 3 points, and power = 3 is used to impute entries. The features for IDW are longitude, latitude, and time of day. To ensure that only the closest coordinates are used, longitude and latitude are over-weighted by 50 times, and only readings from the previous 2 days are used. This ensures that only past values are used at the same time of day from recent days. Note that using other models such as Random Forests causes strong overfitting downstream making IDW ideal. See Figure \ref{fig:feat_eng} for an overview of this pipeline.

\vspace{-5pt}
\subsection{Baselines}
\vspace{-1pt}
The baselines used are common regression models: Ridge Regressor from Scikit-learn \citep{scikit-learn}, XGBoost \citep{chen_xgboost_2016}, LightGBM \citep{LightGBM}, CatBoost \citep{prokhorenkova_catboost_2019}, and a custom inverse distance weighting model optimized using a KDTree. To make a 24-hour forecast, each model uses PM$_{2.5}$ and PM$_{10}$ concentrations 1, 2, 3, and 7 days ago at the same time of day per cell. When any value is not available, the last reading is used. 

\vspace{-5pt}
\subsection{Residual Neural Networks}
\vspace{-2pt}
The next step to model temporal relations is to use residual neural networks and gated recurrent units. RNNs work by maintaining a hidden state $\mathbf{h}_t \in \mathbb{R}^d$ that stores a context of the time series up to time $t$. The final hidden state is used to make the combined forecast using a fully connected layer. $\mathbf{h}_t$ is updated using (\ref{eq:RNN_eqn}) where $\mathbf{W}_{xh}, \mathbf{W}_{hh}$ are weights, $\mathbf{b}_{h}$ is bias and $\mathbf{x}_t$ are the features at time $t$.
\begin{equation}
\mathbf{h}_t = \tanh(\mathbf{W}_{xh} \mathbf{x}_t + \mathbf{W}_{hh} \mathbf{h}_{t-1} + \mathbf{b}_h)
\label{eq:RNN_eqn}
\end{equation}
GRUs have an additional gating mechanism which is used to adaptively choose how much context to retain and how much to absorb from the next timestep. The update is given in (Equation \ref{eq:gru-vs-convgru}). The learnable parameters are all the matrices $\mathbf{W},\mathbf{U}$. Additional model variants are considered that use satellite images that are passed through a Full CNN and concatenated to the final hidden state. This is similar to Figure \ref{fig:conv_gru}, which shows a ConvGRU.

\vspace{-3pt}
\subsection{Spatio-Temporal Graph Neural Networks}

Similar to several modern works in AQI prediction, this project also explores Spatio-temporal Graph Neural Networks. Both graph convolution networks (GCNs) and graph attention networks (GATs) are explored. In a layer of GCN, spatial relations are computed as seen in class:
\vspace{-4pt}
\begin{equation}\mathbf{X}_{t+1} = \sigma\left( \mathring{\mathbf{D}}^{-1/2} \mathring{\mathbf{A}} \mathring{\mathbf{D}}^{-1/2} \mathbf{X}_t \mathbf{W} \right)\end{equation}
Where, $\mathring{\mathbf{A}} = \mathbf{A} + \mathbf{I}$ is the adjacency matrix with added self-loops, $\mathring{\mathbf{D}}$ is the corresponding degree matrix, $\mathbf{W}$ is a learnable weight matrix and $\sigma$ is ReLU, and $\mathbf{X}_t$ represents the feature representation.

GATs learn additional attention weights for each edge describing which neighbors are more useful. In a single layer, a shared weight matrix is applied to every node and using an attention mechanism (a single layer feed-forward neural network parameterized by a vector $\mathbf{a}$ is used) to obtain attention coefficients as in the formula below \citep{velickovic_graph_2018}.

\vspace{-15pt}
\begin{alignat}{2}
\hspace{-1cm}
\textbf{Attention:} \quad
\alpha_{ij} &= 
\frac{
\exp\left( \text{LeakyReLU}\left( \mathbf{a}^\top [\mathbf{W} \mathbf{x}_i \, || \, \mathbf{W} \mathbf{x}_j] \right) \right)
}{
\sum_{k \in \mathcal{N}(i)} \exp\left( \text{LeakyReLU}\left( \mathbf{a}^\top [\mathbf{W} \mathbf{x}_i \, || \, \mathbf{W} \mathbf{x}_k] \right) \right)
}
\quad &
\textbf{, Update:} \quad
\hat{\mathbf{x}}_i &=
\sigma\left( \sum_{j \in \mathcal{N}(i)} \alpha_{ij} \mathbf{W} \mathbf{x}_j \right)
\label{eq:gat}
\end{alignat}

\vspace{-10pt}
The updated features of each node $\hat{\mathbf{x}}_i$ are a weighted mean of a linear transformation of its neighbors through $\mathbf{W}$, with additional non-linearity through $\sigma$ which is ReLU in my implementation.

The STGNN in this report uses a number of GCN or GAT layers followed by a GRU. This models both spatial relations and temporal features. The data processing here is the same as for GRU, with an additional sequence of transformations that convert the tabular sequence data into a sequence of graphs. The graph itself is created by connecting each vertex to its $k$ nearest neighbors. In the experiments later, $k$ ranges from 2 to 7. An example of a graph used is given in Figure \ref{fig:gnn_graph}.

\vspace{-3pt}
\subsection{Convolutional Gated Recurrent Unit}

In order to extend predictions to grid cells where there are no readings, we propose using a Convolutional GRU model. This model is similar to a GRU, with the difference that the inputs $\mathbf{X}_t$ and hidden states $\mathbf{H}_t$ are all 2 dimensional. This allows the model to capture spatial relations from neighbors, which accumulate from across the entire region. The update computations are in (Equation \ref{eq:gru-vs-convgru}). This model introduces stronger inductive bias and translational invariance, intuitively making it better suited to generalize to unseen spatial regions beyond the training area than previous models.

\begin{figure}[htbp]
    \centering
    \includegraphics[width=0.95\textwidth]{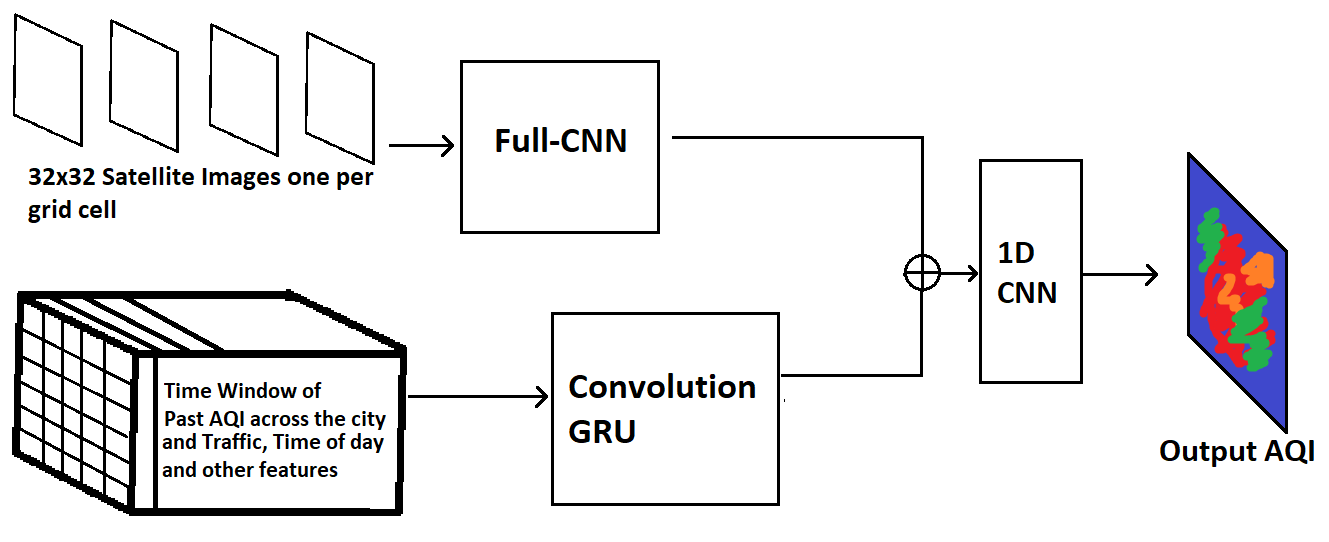}
    \caption{Architecture of the Convolutional GRU proposed in this paper. All satellite image models use a similar pipeline, only swapping out the 1D CNN with a linear layer, and ConvGRU with an RNN, GRU, or STGNN. The Full-CNN architecture is also shared across models.}
    \label{fig:conv_gru}
\end{figure}

The data needs to be converted into a 2D grid, and IDW imputations are used to fill empty cells here. Satellite image features are also used here as for other models. This model is illustrated in Figure \ref{fig:conv_gru}.

\begin{equation}
\begin{tabular}{@{}rl@{\hspace{1cm}}rl@{}}
\multicolumn{2}{c}{\textbf{\hspace{-2cm}Standard GRU}} & \multicolumn{2}{c}{\hspace{-1cm}\textbf{ConvGRU}} \\
$\mathbf{z}_t =$ & $\sigma(\mathbf{W}_z \mathbf{x}_t + \mathbf{U}_z \mathbf{h}_{t-1} + \mathbf{b}_z)$ 
& $\mathbf{Z}_t =$ & $\sigma\left( \mathrm{Conv}_z(\mathbf{X}_t) + \mathrm{Conv}_{hz}(\mathbf{H}_{t-1}) \right)$ \\
$\mathbf{r}_t =$ & $\sigma(\mathbf{W}_r \mathbf{x}_t + \mathbf{U}_r \mathbf{h}_{t-1} + \mathbf{b}_r)$ 
& $\mathbf{R}_t =$ & $\sigma\left( \mathrm{Conv}_r(\mathbf{X}_t) + \mathrm{Conv}_{hr}(\mathbf{H}_{t-1}) \right)$ \\
$\tilde{\mathbf{h}}_t =$ & $\tanh(\mathbf{W}_h \mathbf{x}_t + \mathbf{U}_h (\mathbf{r}_t \odot \mathbf{h}_{t-1}) + \mathbf{b}_h)$ 
& $\tilde{\mathbf{H}}_t =$ & $\tanh\left( \mathrm{Conv}_h(\mathbf{X}_t) + \mathrm{Conv}_{hh}(\mathbf{R}_t \odot \mathbf{H}_{t-1}) \right)$ \\
$\mathbf{h}_t =$ & $(1 - \mathbf{z}_t) \odot \mathbf{h}_{t-1} + \mathbf{z}_t \odot \tilde{\mathbf{h}}_t$ 
& $\mathbf{H}_t =$ & $(1 - \mathbf{Z}_t) \odot \mathbf{H}_{t-1} + \mathbf{Z}_t \odot \tilde{\mathbf{H}}_t$
\end{tabular}
\label{eq:gru-vs-convgru}
\end{equation}

\vspace{-5pt}
\section{Main Results}

By doing a sweep on the number of RNN/GRU layers of each model, we obtain Table \ref{tab:model_results} of the best 24-hour forecast results for AirDelhi. It displays the RMSE, $R^2$ score, and Accuracy. The first observation is that the models proposed in this paper greatly outperform the baseline models.

\begin{table}[h!]
\centering
\caption{Model Comparison on Test and Extended Test Datasets (Best Results Highlighted)}
\label{tab:model_results}
\hspace*{-2cm}
\begin{tabular}{lccc|ccc|ccc|ccc}
\toprule
\textbf{Model} & \multicolumn{6}{c|}{\textbf{PM\textsubscript{2.5}}} & \multicolumn{6}{c}{\textbf{PM\textsubscript{10}}} \\
 & \multicolumn{3}{c|}{Test} & \multicolumn{3}{c|}{Extended} & \multicolumn{3}{c|}{Test} & \multicolumn{3}{c}{Extended} \\
 & RMSE & $R^2$ & Acc\% & RMSE & $R^2$ & Acc\% & RMSE & $R^2$ & Acc\% & RMSE & $R^2$ & Acc\% \\
\midrule
XGB & 98.1 & 0.053 & 50.8 & 97 & 0.081 & 51.4 & 106 & 0.066 & 52.8 & 103 & 0.115 & 53.65  \\
Ridge & 90.6 & 0.192 & 56.2 & 90.9 & 0.193 & 55.9 & 98.5 & 0.197 & 56 & 98.1 & 0.196 & 56.39  \\
LightGBM & 90.8 & 0.190 & 55 & 90.7 & 0.197 & 54.8 & 98.7 & 0.194 & 56.1 & 98.5 & 0.191 & 55.71  \\
CatBoost & 94.9 & 0.114 & 53 & 95.1 & 0.118 & 52.9 & 102 & 0.130 & 54 & 102 & 0.135 & 54.23  \\
IDW & 92.2 & 0.165 & 55.9 & 92.4 & 0.166 & 55.7 & 100 & 0.165 & 55.1 & 100 & 0.163 & 55.34  \\
\hline
GRU & 20.9 & 0.893 & 98.6 & \highlight{20.6} & \highlight{0.895} & \highlight{98.6} & 20 & 0.891 & 97.9 & \highlight{19.5} & \highlight{0.895} & 98 \\
RNN & 21.8 & 0.885 & 98.4 & 21.7 & 0.884 & 98.4 & 20.7 & 0.883 & 97.8 & 20.3 & 0.886 & 97.9 \\
GRU Image & 21.1 & 0.891 & 98.5 & 20.7 & 0.893 & \highlight{98.6} & 19.9 & 0.892 & \highlight{98.1} & 19.6 & 0.894 & \highlight{98.1} \\
RNN Image & 22.1 & 0.881 & 98.3 & 21.6 & 0.884 & 98.5 & 20.6 & 0.885 & 97.9 & 20.1 & 0.888 & 97.9 \\
\hline
GCN & 19.3 & 0.909 & 98.7 & 31.5 & 0.753 & 98 & 18.2 & 0.91 & 97.9 & 29 & 0.767 & 96.8 \\
GAT & 19.1 & 0.911 & 98.7 & 29.2 & 0.789 & 98 & 18.1 & 0.91 & 98 & 27 & 0.799 & 96.9 \\
GCN Image & \highlight{18.9} & \highlight{0.913} & \highlight{98.8} & 24.9 & 0.846 & 98.1 & \highlight{17.8} & \highlight{0.914} & \highlight{98.1} & 23.4 & 0.849 & 97.4 \\
GAT Image & 19.1 & 0.911 & 98.7 & 28.3 & 0.802 & 98 & 18.2 & 0.91 & 98 & 26.4 & 0.807 & 96.9 \\
\hline
CONV & 25.6 & 0.836 & 97.5 & 27.8 & 0.804 & 97.4 & 23.8 & 0.842 & 97.6 & 26.1 & 0.807 & 97.6 \\
\bottomrule
\end{tabular}
\end{table}

It can also be observed that GRUs outperform RNNs by 0.01 $R^2$ score on both responders (justifying using GRUs as a base for more complex models). This result is expected because RNNs usually perform worse than GRUs on long-sequence problems (105 steps including forecasting). Satellite images also only serve as a noise feature for GRUs and RNNs. We note that when evaluating on unseen coordinates, the performance improves (columns titled "Full Test"). This suggests that these models are not memorizing coordinates, and that there are strong temporal patterns in the two-day input time window that are coordinate agnostic, learnable, and repeatable.

On the other hand, STGNN-based approaches outperform GRUs and RNNs by 0.02 R2 score. This improvement is not as large as expected because using IDW for imputation pools measurements at the same time from nearby coordinates which provide some spatial information for GRUs and RNNs as well. Thus, STGNNs only provide minimal advantage over RNNs. However, STGNNs overfit on spatial relations because adding more (unseen) nodes decreases R2 score by up to 0.15. This implies that these STGNNs are not strong at adjusting to new graphs. However, using satellite images for STGNNs reduces this degradation for GATs by 0.02 R2 and improves test performance slightly.

The Convolutional GRU model is unable to learn as well as other models because for each timestep it only has 4\% of the spatial nodes as true labels. This sparsity limits the model's ability to learn temporal dependencies across the spatial domain. It's performance doesn't degrade significantly on distant unseen coordinates, demonstrating that this model is able to extend forecasts to unseen regions with a performance greatly exceeding any baseline. 

For classifying AQI labels, the performance of all the proposed models is similar and is significantly higher than the baselines. Mostly, it aligns with regression objectives, except for GRU Image which achieves higher accuracy with lower $R^2$ in the Extended test set and PM$_{10}$ test set.

\vspace{-2pt}
\subsection{Ablation and Explainability Studies}

This section focuses on sensitivity to various hyperparameters, and inspecting the ability of the designed models to forecast for longer periods of time. Since the performance of models with Satellite images is similar to those without, only standard models are inspected for conciseness. Note that for models using a Graph or 2D representation, the dataset size is fewer than 70 datapoints, which can make models more prone to overfitting and instability during training. 

\vspace{-2pt}
\subsubsection{Model Size}

\begin{figure}[htbp]
    \centering
    \begin{subfigure}[t]{0.48\textwidth}
        \centering
        \includegraphics[width=\textwidth]{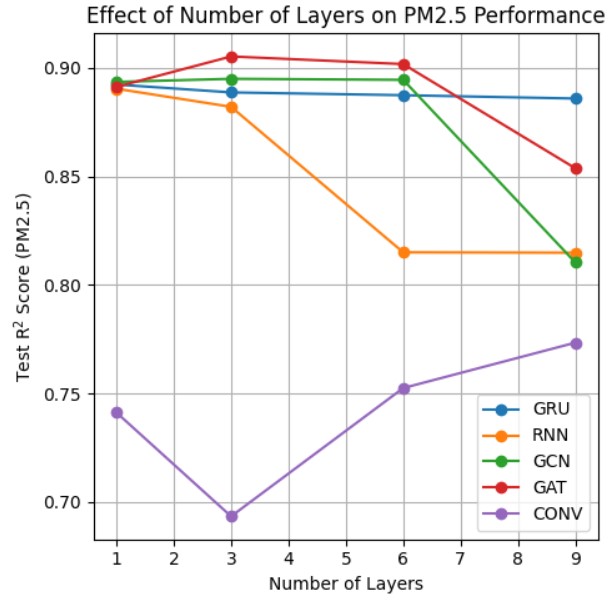}
        \caption{PM$_{2.5}$}
        \label{fig:model_size_pm25}
    \end{subfigure}
    \hfill
    \begin{subfigure}[t]{0.48\textwidth}
        \centering
        \includegraphics[width=\textwidth]{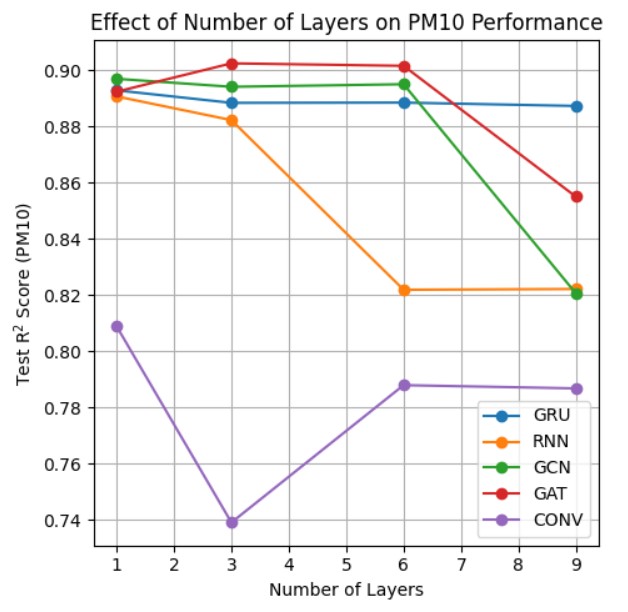}
        \caption{PM 10}
        \label{fig:model_size_pm10}
    \end{subfigure}
    \caption{Test $R^2$ score for PM$_{2.5}$ and PM$_{10}$ as model size changes. GRU has a stable performance, and ConvGRU starts improving, whereas other models begin overfitting.}
    \label{fig:ablation_model_size}
\end{figure}

One of the most important parameters while creating a model would be the model's size. In Figure \ref{fig:ablation_model_size} we can see that the performance of most models is optimal approximately around 3 or 6 layers, after which it begins to overfit (except for ConvGRU which performs better with more layers - further suggesting that it might not have sufficient capacity). Note that the plain GRU is very stable to model size (its performance does not change).

\vspace{-2pt}
\subsubsection{Sequence Length}

To analyze the impact of temporal features, we can analyze the effect of sequence length on model performance for forecasting the next day. For this experiment, all models forecast the next 24 hours and have the same number of layers and size. 

\begin{figure}[htbp]
    \centering
    \begin{subfigure}[t]{0.48\textwidth}
        \centering
        \includegraphics[width=\textwidth]{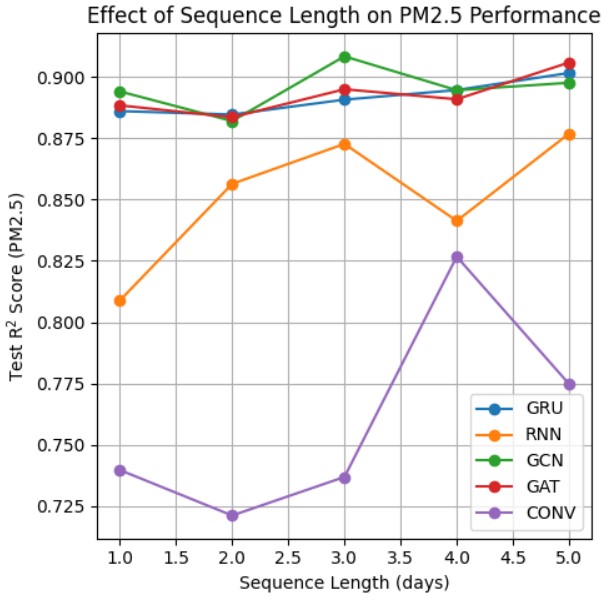}
        \label{fig:seq_length_pm25}
    \end{subfigure}
    \hfill
    \begin{subfigure}[t]{0.48\textwidth}
        \centering
        \includegraphics[width=\textwidth]{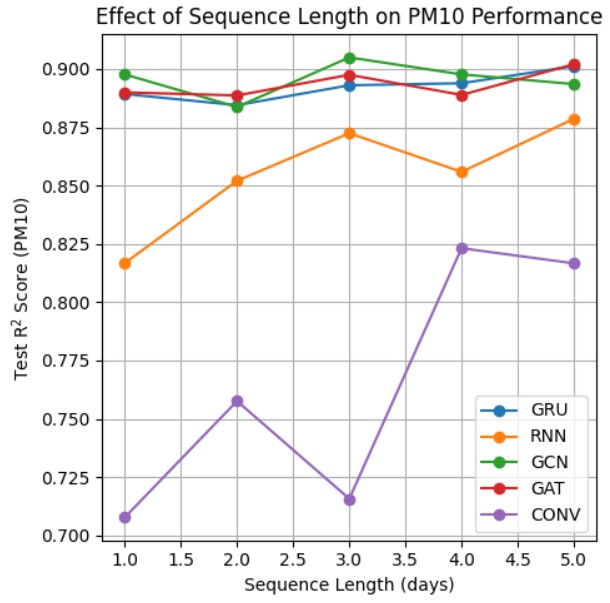}
        \label{fig:seq_length_pm10}
    \end{subfigure}
    \caption{Test $R^2$ score for PM$_{2.5}$ and PM$_{10}$ vs input sequence length. Generally, performance improves slightly as longer sequence is input, aside from a few outliers due to randomness in training}
    \label{fig:ablation_seq_len}
\end{figure}

If we expect a long-term trend to appear that would help predictions, the longer the context window, the better the performance should be. However, for AirDelhi, \citet{AirDelhi} observe strong correlations between each subsequent hour. This suggests that using longer time windows has minor gains. In Figure \ref{fig:ablation_seq_len} we observe that in general, a longer sequence improves performance. However, we see only minor improvements (roughly 0.015 R2 score for GRU, GCN, GAT, but major improvements for RNN and ConvGRU, ignoring the aforementioned training noise) - supporting our reasoning. Nevertheless, there is some gain if the input time window is extended. 

\subsubsection{Forecast Horizon}

An important parameter to estimate is also the forecast horizon: how far ahead can these models predict reliably. For this experiment, models have the same size (same number of layers, hidden dimension), the same input sequence length and use satellite images.

\begin{figure}[htbp]
    \centering
    \begin{subfigure}[t]{0.45\textwidth}
        \centering
        \includegraphics[width=\textwidth]{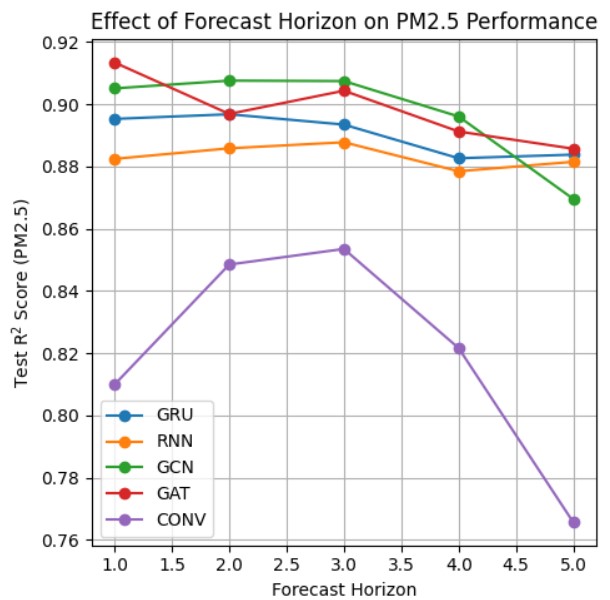}
        \label{fig:horizon_pm25}
    \end{subfigure}
    \hfill
    \begin{subfigure}[t]{0.45\textwidth}
        \centering
        \includegraphics[width=\textwidth]{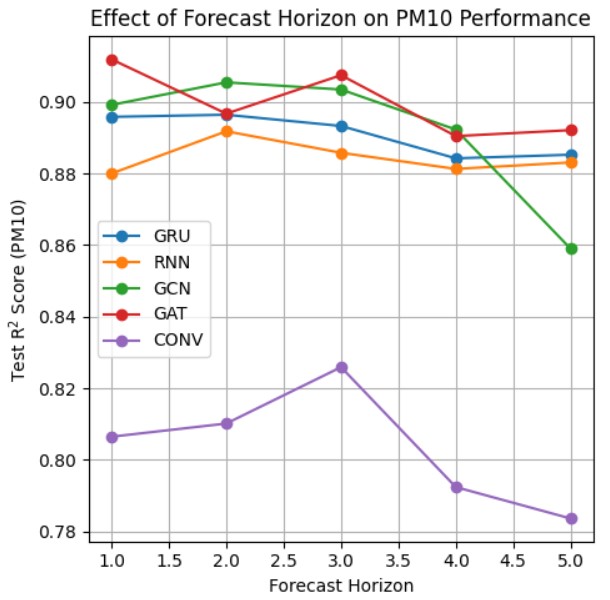}
        \label{fig:horizon_pm10}
    \end{subfigure}
    \caption{Test $R^2$ score for PM$_{2.5}$ and PM$_{10}$ vs forecast horizon. Generally, performance decreases slightly for longer horizons, but the drop is not large}
    \label{fig:ablation_horizon}
\end{figure}

Figure \ref{fig:ablation_horizon} shows that a forecast horizon of 5 days has a performance drop for GRUs, STGNNs of ~0.01 to 0.04 $R^2$, which is not large. This suggests that the RNN, GRU and STGNN based models are able to gain enough contextual information to predict the next few days reliably. This suggests that AQI typically demonstrates recurring patterns learnable by recurrent models. As baselines receive sparsely lagged features they cannot learn these patterns and thus perform poorly.

$R^2$ for ConvGRU rises then sharply falls. Initially, as longer forecast samples contain more true labels, gradients are more reliable. However, for longer times, performance falls like other models.

\vspace{-2pt}
\subsubsection{Number of Neighbors}

\begin{figure}[htbp]
    \centering
    \begin{subfigure}[t]{0.45\textwidth}
        \centering
        \includegraphics[width=\textwidth]{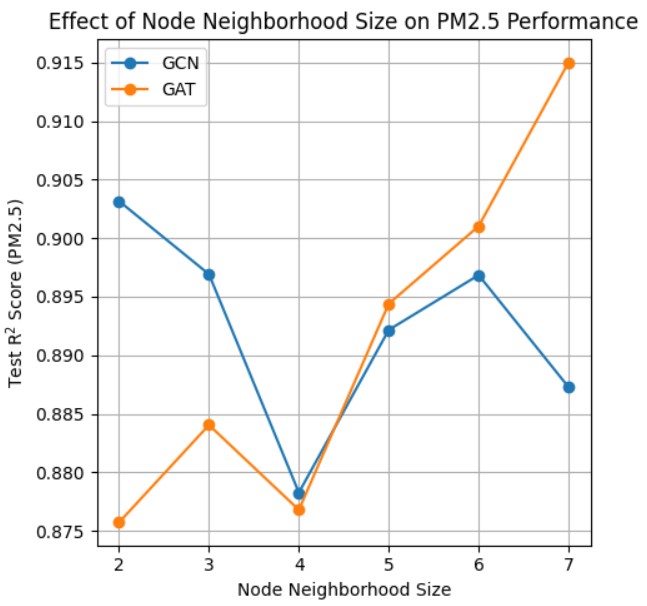}
        \label{fig:node_pm25}
    \end{subfigure}
    \hfill
    \begin{subfigure}[t]{0.45\textwidth}
        \centering
        \includegraphics[width=\textwidth]{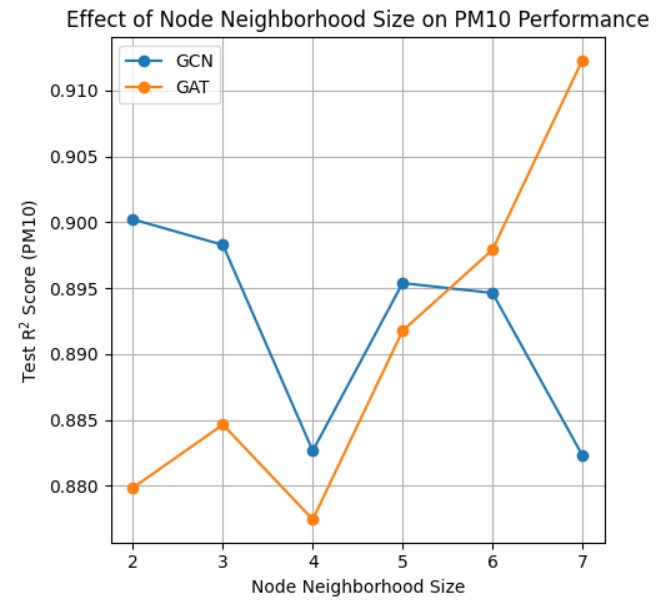}
        \label{fig:node_pm10}
    \end{subfigure}
    \caption{Test $R^2$ score for PM$_{2.5}$ and PM$_{10}$ vs node degree. While GCNs overfit with more neighbors, GATs improve performance by attending to correct neighbors.}
    \label{fig:ablation_nodes}
\end{figure}

STGNNs allow us to also investigate the impact of spatial relations by looking at the number of neighbors of each node. GCN-based models perform worse with more neighbors, but GAT-based models can identify important neighbors using attention, and only use good signals thus improving $R^2$ as vertex degree increases (Figure \ref{fig:ablation_nodes}). This suggests that there are spatial relations between adjacent grid cells vary across time, otherwise, GCN performance would remain constant or improve as it would get similar signals from neighbors in both train and test. This is likely due to changing weather patterns (wind, atmospheric pressure) causing some neighbors to be more important at different times. With more neighbors, GAT achieves the highest PM$_{2.5}$ $R^2$ score of 0.915.


\vspace{-5pt}
\section{Conclusion}
\vspace{-5pt}

This report has presented the first in-depth modeling approach to forecast AQI over New Delhi at a fine-grained level by presenting several models capable of forecasting AQI for a period of 5 days. The models presented significantly surpass existing baselines, thus providing a reliable pipeline for fine-grained AQI prediction even to completely unseen coordinates. This report also uncovers new insights into air pollution in New Delhi, such as recurring short-term patterns, strong short-term temporal correlation, and presence of changing spatial relations.

Due to resource limitations, many important features could not be included in this study. It is likely that additional weather features, such as wind speed, pressure, real-time traffic, neighborhood description, population and other features can improve performance further. Alternative imputation methods such as diffusion and GANs, which could be more realistic could  also be explored. 

Another direction is to collect a larger dataset than AirDelhi (which only spans 3 months), to instead span more than a year and include more cities. This dataset should consider other pollutants such as NO$_x$ and SO$_x$. This increased data would be useful for building better models and be more general. Testing models trained on AirDelhi in other cities (on sparse datasets) would be another extension. 

\newpage

\section{Acknowledgement}
I would like to thank my mother for giving me inspiration for this topic. I would like to thank Professor Yaoliang Yu who constantly guided me and answered all my questions after every class in CS 480!


\begin{thebibliography}{99}

\providecommand{\natexlab}[1]{#1}
\providecommand{\url}[1]{\texttt{#1}}
\expandafter\ifx\csname urlstyle\endcsname\relax
  \providecommand{\doi}[1]{doi: #1}
\else
  \providecommand{\doi}{doi: \begingroup \urlstyle{rm}\Url}
\fi


\bibitem[Chen(2025)]{chen_air_2025}
Ying Chen.
\newblock Air pollution in {New} {Delhi} is more severe than observed due to hygroscopicity-induced bias in aerosol sampling.
\newblock \emph{npj Clean Air}, 1\penalty0 (1):\penalty0 1, March 2025.
\newblock ISSN 3059-2240.
\newblock \doi{10.1038/s44407-024-00001-6}.
\newblock URL \url{https://www.nature.com/articles/s44407-024-00001-6}.

\bibitem[Bank(2024)]{WorldBank}
World Bank.
\newblock How is india trying to address air pollution?, 2024.
\newblock URL
  \url{https://www.worldbank.org/en/country/india/publication/catalyzing-clean-air-in-india?}

\bibitem[Sahu et~al.(2020)Sahu, Dixit, Mishra, Kumar, Shukla, Sutaria, Tiwari, and Tripathi]{sahu_validation_2020}
Ravi Sahu, Kuldeep~Kumar Dixit, Suneeti Mishra, Purushottam Kumar, Ashutosh~Kumar Shukla, Ronak Sutaria, Shashi Tiwari, and Sachchida~Nand Tripathi.
\newblock Validation of {Low}-{Cost} {Sensors} in {Measuring} {Real}-{Time} {PM10} {Concentrations} at {Two} {Sites} in {Delhi} {National} {Capital} {Region}.
\newblock \emph{Sensors}, 20\penalty0 (5):\penalty0 1347, February 2020.
\newblock ISSN 1424-8220.
\newblock \doi{10.3390/s20051347}.
\newblock URL \url{https://www.mdpi.com/1424-8220/20/5/1347}.

\bibitem[Chauhan et~al.(2023)Chauhan, Patel, Ranu, Sen, and Batra]{AirDelhi}
Sachin Chauhan, Zeel~Bharatkumar Patel, Sayan Ranu, Rijurekha Sen, and Nipun Batra.
\newblock {AirDelhi: Fine-Grained Spatio-Temporal Particulate Matter Dataset from Delhi for ML-Based Modeling}.
\newblock In A.~Oh, T.~Naumann, A.~Globerson, K.~Saenko, M.~Hardt, and S.~Levine, editors, \emph{Advances in Neural Information Processing Systems}, volume~36, pages 75455--75468. Curran Associates, Inc., 2023.
\newblock URL \url{https://proceedings.neurips.cc/paper_files/paper/2023/file/ee799aff607fcf39c01df6391e96f92c-Paper-Datasets_and_Benchmarks.pdf}.

\bibitem[Rad et~al.(2025)Rad, Nematollahi, Pak, and Mahmoudi]{rad_predictive_2025}
Abdullah~Kaviani Rad, Mohammad~Javad Nematollahi, Abbas Pak, and Mohammadreza Mahmoudi.
\newblock Predictive modeling of air quality in the {Tehran} megacity via deep learning techniques.
\newblock \emph{Scientific Reports}, 15\penalty0 (1):\penalty0 1367, January 2025.
\newblock ISSN 2045-2322.
\newblock \doi{10.1038/s41598-024-84550-6}.
\newblock URL \url{https://www.nature.com/articles/s41598-024-84550-6}.

\bibitem[Pruthi and Liu(2022)]{pruthi_low-cost_2022}
D.~Pruthi and Y.~Liu.
\newblock Low-cost nature-inspired deep learning system for {PM2}.5 forecast over {Delhi}, {India}.
\newblock \emph{Environment International}, 166:\penalty0 107373, August 2022.
\newblock ISSN 01604120.
\newblock \doi{10.1016/j.envint.2022.107373}.
\newblock URL \url{https://linkinghub.elsevier.com/retrieve/pii/S0160412022003002}.

\bibitem[Alawi et~al.(2024)Alawi, Kamar, Alsuwaiyan, and Yaseen]{alawi_temporal_2024}
Omer~A. Alawi, Haslinda~Mohamed Kamar, Ali Alsuwaiyan, and Zaher~Mundher Yaseen.
\newblock Temporal trends and predictive modeling of air pollutants in {Delhi}: a comparative study of artificial intelligence models.
\newblock \emph{Scientific Reports}, 14\penalty0 (1):\penalty0 30957, December 2024.
\newblock ISSN 2045-2322.
\newblock \doi{10.1038/s41598-024-82117-z}.
\newblock URL \url{https://www.nature.com/articles/s41598-024-82117-z}.

\bibitem[Méndez et~al.(2023)Méndez, Merayo, and Núñez]{mendez_machine_2023}
Manuel Méndez, Mercedes~G. Merayo, and Manuel Núñez.
\newblock Machine learning algorithms to forecast air quality: a survey.
\newblock \emph{Artificial Intelligence Review}, 56\penalty0 (9):\penalty0 10031--10066, September 2023.
\newblock ISSN 0269-2821, 1573-7462.
\newblock \doi{10.1007/s10462-023-10424-4}.
\newblock URL \url{https://link.springer.com/10.1007/s10462-023-10424-4}.

\bibitem[Parameshachari et~al.(2022)Parameshachari, Siddesh, Sridhar, Latha, Sattar, and Manjula.]{parameshachari_prediction_2022}
B~D Parameshachari, G~M Siddesh, V.~Sridhar, M~Latha, Khalid Nazim~Abdul Sattar, and G~Manjula.
\newblock Prediction and {Analysis} of {Air} {Quality} {Index} using {Machine} {Learning} {Algorithms}.
\newblock In \emph{2022 {IEEE} {International} {Conference} on {Data} {Science} and {Information} {System} ({ICDSIS})}, pages 1--5, Hassan, India, July 2022. IEEE.
\newblock ISBN 978-1-66549-801-2.
\newblock \doi{10.1109/ICDSIS55133.2022.9915802}.
\newblock URL \url{https://ieeexplore.ieee.org/document/9915802/}.

\bibitem[Hettige et~al.(2024)Hettige, Ji, Xiang, Long, Cong, and Wang]{hettige_airphynet_2024}
Kethmi~Hirushini Hettige, Jiahao Ji, Shili Xiang, Cheng Long, Gao Cong, and Jingyuan Wang.
\newblock {AirPhyNet}: {Harnessing} {Physics}-{Guided} {Neural} {Networks} for {Air} {Quality} {Prediction}, February 2024.
\newblock URL \url{http://arxiv.org/abs/2402.03784}.
\newblock arXiv:2402.03784 [cs].

\bibitem[Guan et~al.(2020)Guan, Johnson, Katzfuss, Mannshardt, Messier, Reich, and Song]{guan_fine-scale_2020}
Yawen Guan, Margaret Johnson, Matthias Katzfuss, Elizabeth Mannshardt, Kyle~P. Messier, Brian~J. Reich, and Joon~Jin Song.
\newblock Fine-scale spatiotemporal air pollution analysis using mobile monitors on {Google} {Street} {View} vehicles.
\newblock \emph{Journal of the American Statistical Association}, 115\penalty0 (531):\penalty0 1111--1124, July 2020.
\newblock ISSN 0162-1459, 1537-274X.
\newblock \doi{10.1080/01621459.2019.1665526}.
\newblock URL \url{http://arxiv.org/abs/1810.03576}.
\newblock arXiv:1810.03576 [stat].

\bibitem[Zhao et~al.(2023)Zhao, Fan, Xia, Jia, Wang, and Yang]{zhao_astgc_2023}
Yaning Zhao, Shurui Fan, Kewen Xia, Yingmiao Jia, Li~Wang, and Wenbiao Yang.
\newblock {ASTGC}: {Attention}-based {Spatio}-temporal {Fusion} {Graph} {Convolution} {Model} for {Fine}-grained {Air} {Quality} {Analysis}.
\newblock \emph{Air Quality, Atmosphere \& Health}, 16\penalty0 (9):\penalty0 1761--1775, September 2023.
\newblock ISSN 1873-9318, 1873-9326.
\newblock \doi{10.1007/s11869-023-01369-2}.
\newblock URL \url{https://link.springer.com/10.1007/s11869-023-01369-2}.

\bibitem[Rowley and Karakuş(2023)]{rowley_predicting_2023}
Andrew Rowley and Oktay Karakuş.
\newblock Predicting air quality via multimodal {AI} and satellite imagery.
\newblock \emph{Remote Sensing of Environment}, 293:\penalty0 113609, August 2023.
\newblock ISSN 00344257.
\newblock \doi{10.1016/j.rse.2023.113609}.
\newblock URL \url{http://arxiv.org/abs/2211.00780}.
\newblock arXiv:2211.00780 [cs].

\bibitem[Song and Han(2020)]{song_deep-maps_2020}
Jun Song and Ke~Han.
\newblock Deep-{MAPS}: {Machine} {Learning} based {Mobile} {Air} {Pollution} {Sensing}, March 2020.
\newblock URL \url{http://arxiv.org/abs/1904.12303}.
\newblock arXiv:1904.12303 [cs].

\bibitem[Nguyen et~al.(2024)Nguyen, Pham, Oo, Ahn, and Lim]{nguyen_predicting_2024}
Anh~Tuan Nguyen, Duy~Hoang Pham, Bee~Lan Oo, Yonghan Ahn, and Benson T.~H. Lim.
\newblock Predicting air quality index using attention hybrid deep learning and quantum-inspired particle swarm optimization.
\newblock \emph{Journal of Big Data}, 11\penalty0 (1):\penalty0 71, May 2024.
\newblock ISSN 2196-1115.
\newblock \doi{10.1186/s40537-024-00926-5}.
\newblock URL \url{https://journalofbigdata.springeropen.com/articles/10.1186/s40537-024-00926-5}.

\bibitem[Natarajan et~al.(2024)Natarajan, Shanmurthy, Arockiam, Balusamy, and Selvarajan]{natarajan_optimized_2024}
Suresh~Kumar Natarajan, Prakash Shanmurthy, Daniel Arockiam, Balamurugan Balusamy, and Shitharth Selvarajan.
\newblock Optimized machine learning model for air quality index prediction in major cities in {India}.
\newblock \emph{Scientific Reports}, 14\penalty0 (1):\penalty0 6795, March 2024.
\newblock ISSN 2045-2322.
\newblock \doi{10.1038/s41598-024-54807-1}.
\newblock URL \url{https://www.nature.com/articles/s41598-024-54807-1}.

\bibitem[India(2025)]{India_AQI}
Govt. India.
\newblock Central pollution control board, Apr 2025.
\newblock URL \url{https://cpcb.nic.in/National-Air-Quality-Index/}.

\bibitem[Pedregosa et~al.(2011)Pedregosa, Varoquaux, Gramfort, Michel, Thirion, Grisel, Blondel, Prettenhofer, Weiss, Dubourg, Vanderplas, Passos, Cournapeau, Brucher, Perrot, and Duchesnay]{scikit-learn}
F.~Pedregosa, G.~Varoquaux, A.~Gramfort, V.~Michel, B.~Thirion, O.~Grisel, M.~Blondel, P.~Prettenhofer, R.~Weiss, V.~Dubourg, J.~Vanderplas, A.~Passos, D.~Cournapeau, M.~Brucher, M.~Perrot, and E.~Duchesnay.
\newblock Scikit-learn: Machine learning in {P}ython.
\newblock \emph{Journal of Machine Learning Research}, 12:\penalty0 2825--2830, 2011.

\bibitem[Chen and Guestrin(2016)]{chen_xgboost_2016}
Tianqi Chen and Carlos Guestrin.
\newblock {XGBoost}: {A} {Scalable} {Tree} {Boosting} {System}.
\newblock In \emph{Proceedings of the 22nd {ACM} {SIGKDD} {International} {Conference} on {Knowledge} {Discovery} and {Data} {Mining}}, pages 785--794, San Francisco California USA, August 2016. ACM.
\newblock ISBN 978-1-4503-4232-2.
\newblock \doi{10.1145/2939672.2939785}.
\newblock URL \url{https://dl.acm.org/doi/10.1145/2939672.2939785}.

\bibitem[Ke et~al.(2017)Ke, Meng, Finley, Wang, Chen, Ma, Ye, and
  Liu]{LightGBM}
Guolin Ke, Qi~Meng, Thomas Finley, Taifeng Wang, Wei Chen, Weidong Ma, Qiwei
  Ye, and Tie-Yan Liu.
\newblock Lightgbm: A highly efficient gradient boosting decision tree.
\newblock In I.~Guyon, U.~Von Luxburg, S.~Bengio, H.~Wallach, R.~Fergus,
  S.~Vishwanathan, and R.~Garnett, editors, \emph{Advances in Neural
  Information Processing Systems}, volume~30. Curran Associates, Inc., 2017.
\newblock URL
  \url{https://proceedings.neurips.cc/paper_files/paper/2017/file/6449f44a102fde848669bdd9eb6b76fa-Paper.pdf}.

\bibitem[Prokhorenkova et~al.(2019)Prokhorenkova, Gusev, Vorobev, Dorogush, and Gulin]{prokhorenkova_catboost_2019}
Liudmila Prokhorenkova, Gleb Gusev, Aleksandr Vorobev, Anna~Veronika Dorogush, and Andrey Gulin.
\newblock {CatBoost}: unbiased boosting with categorical features, January 2019.
\newblock URL \url{http://arxiv.org/abs/1706.09516}.
\newblock arXiv:1706.09516 [cs].

\bibitem[Veličković et~al.(2018)Veličković, Cucurull, Casanova, Romero, Liò, and Bengio]{velickovic_graph_2018}
Petar Veličković, Guillem Cucurull, Arantxa Casanova, Adriana Romero, Pietro Liò, and Yoshua Bengio.
\newblock Graph {Attention} {Networks}, February 2018.
\newblock URL \url{http://arxiv.org/abs/1710.10903}.
\newblock arXiv:1710.10903 [stat].

\end{thebibliography}
\end{document}